\documentclass[10pt,twocolumn ,letterpaper]{article}

\usepackage{cvpr}
\usepackage{times}
\usepackage{epsfig}
\usepackage{graphicx}
\usepackage{amsmath}
\usepackage{amssymb}

\usepackage{comment}
\usepackage{bm}

\usepackage[pagebackref=true,breaklinks=true,letterpaper=true,colorlinks,bookmarks=false]{hyperref}

\cvprfinalcopy 

\def\cvprPaperID{4406} 

\ifcvprfinal\fi

\title{Rare Event Detection using Disentangled Representation Learning}

\author{Ryuhei Hamaguchi, Ken Sakurada, and Ryosuke Nakamura\\
National Institute of Advanced Industrial Science and Technology (AIST)\\
{\tt\small \{ryuhei.hamaguchi, k.sakurada, r.nakamura\}@aist.go.jp}
}

\begin{document}

\maketitle

\begin{abstract}
This paper presents a novel method for rare event detection from an image pair with class-imbalanced datasets. A straightforward approach for event detection tasks is to train a detection network from a large-scale dataset in an end-to-end manner.
However, in many applications such as building change detection on satellite images, few positive samples are available for the training.
Moreover, scene image pairs contain many trivial events, such as in illumination changes or background motions. 
These many trivial events and the class imbalance problem lead to false alarms for rare event detection.
In order to overcome these difficulties, we propose a novel method to learn disentangled representations from only low-cost negative samples. 
The proposed method disentangles different aspects in a pair of observations: variant and invariant factors that represent trivial events and image contents, respectively. 
The effectiveness of the proposed approach is verified by the quantitative evaluations on four change detection datasets, and the qualitative analysis shows that the proposed method can acquire the representations that disentangle rare events from trivial ones.
\end{abstract}

\section{Introduction}
\label{sect:introduction}
In the field of computer vision, event detection from an image pair has been comprehensively studied as image similarity estimation.
Similarity estimation between images is one of the fundamental problems, which can be applied for many tasks, such as change detection \cite{fujita2017damage,khan2017learning,sakurada2015change,BMVC2015_127}, image retrieval and matching \cite{babenko2014neural,simo2015discriminative,zagoruyko2015learning}, identification \cite{NIPS2014_5416,wen2016discriminative}, and stereo matching \cite{chen2015deep,zbontar2016stereo}. Thanks to the recent success of deep features , the image comparison methods have substantially progressed. However, a general draw back is that they require a large amount of dataset to fully utilize the representational power of the deep features. 

In the context of image similarity estimation, this paper considers a particular task of detecting rare events from an image pair, such as detecting building changes on a pair of satellite images, or detecting manufacturing defects by comparing images of products. One challenge of the task lies in the difficulty of collecting training samples. Because finding rare samples is labor intensive task, the training dataset often includes few positive samples. Additionally, image pairs often contain many cumbersome events that are not of interest ({\it e.g.,} illumination changes, registration error of images, shadow changes, background motion, or seasonal changes). 
These many trivial events and the class imbalance problem lead to false alarms for trivial events, or overlooking the rare events.

\begin{figure}
\centering
\includegraphics[width=8cm]{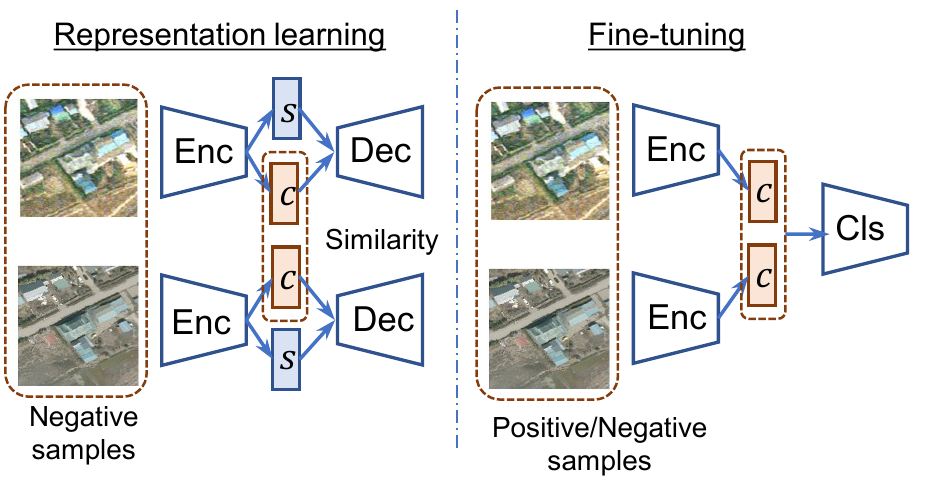}
\caption{The overall concept of the proposed model. From the negative image pairs, the representation learning model ({\bf left}) learns features that are invariant to trivial events. The rare event detector ({\bf right}) is then trained on the learned invariant features.}
\label{fig:concept}
\end{figure}

In order to overcome these difficulties, we propose a novel network architecture for disentangled representation learning using only low-cost negative image pairs. Figure~\ref{fig:concept} demonstrates the overall concept of the proposed method. 
The proposed network is trained to encode each image into the two separated features, {\it specific} and {\it common}, by introducing a similarity constraint between the image contents.
The specific and common features represent a mixture of information related to trivial events ({\it e.g.,} illumination, shadows, or background motion) and image contents that is invariant to trivial events, respectively.
This disentanglement can be learned using only low-cost negative samples because negative samples contain many trivial events. 

The effectiveness of the proposed method on the class-imbalance scenario is verified by the qualitative evaluations on four change detection datasets, including in-the-wild datasets. Qualitative analysis shows that the proposed method successfully learns the disentangled representation for both rare events and trivial ones in the image pairs. The contributions of this work are as follows:
\begin{itemize}
\item We propose a novel solution to the class imbalance problem in rare event detection tasks, which has not been fully studied in the past literature.
\item We propose a novel representation learning method that only requires pairs of observations to learn disentangled representations.
\item We create a new large-scale change detection dataset from the open data repository of Washington D.C.
\end{itemize}

\begin{figure*}
\centering
\includegraphics[width=12cm]{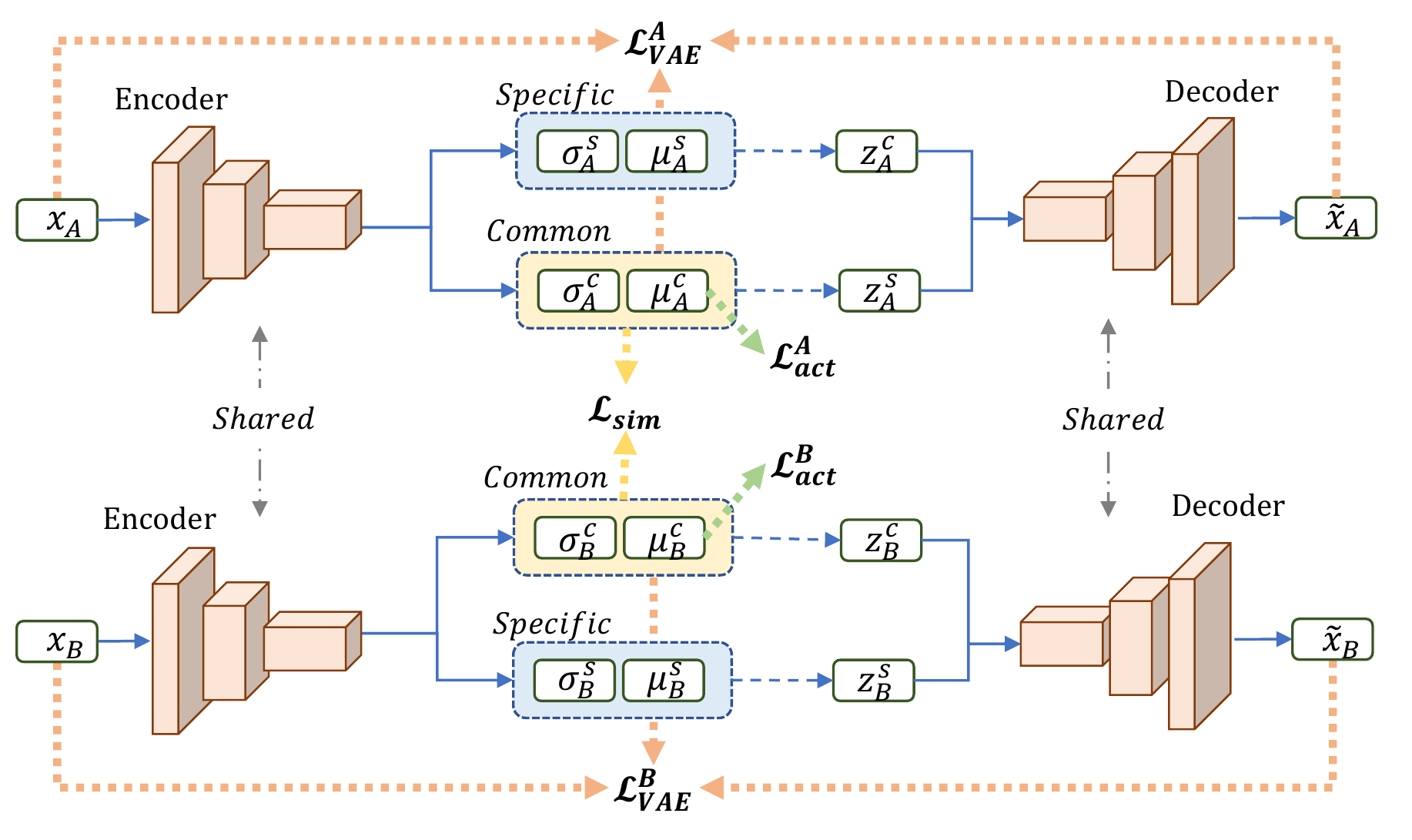}
\caption{Schematics of the proposed representation learning method. The model takes a pair of images $\bm{x}_A$ and $\bm{x}_B$ as input. For each image, the encoder extracts common and specific features, and the decoder reconstructs the input. The key feature of the model is the similarity loss $\mathcal{L}_{sim}$. This loss constrains the common features to extract invariant factors between $\bm{x}_A$ and $\bm{x}_B$. Another feature is the activation loss $\mathcal{L}_{act}$. This loss encourages the mean vector of the common features ($\bm{\mu}^{c}$) to be activated, which avoids a trivial solution -- $(\bm{\sigma}^{c},\bm{\mu}^{c})=(\bm{1},\bm{0})$ -- for any input.}
\label{fig:method_rep_learning}
\end{figure*}


\section{Related Work}
\label{sect:related works}
In change detection tasks, several works have attempted to overcome the difficulties of data collection and cumbersome trivial events as described in the previous section. In order to save the cost of annotation, \cite{Khan2017} proposed a weakly supervised method that requires only image-level labels to train their change segmentation models. Although their work saves the pixel-level annotation cost, it still requires image-level labels, which are difficult to collect for rare change events. To address trivial events, several works on video surveillance tasks \cite{Barnich2011,Stauffer1999} utilize background modeling techniques in which foreground changes are detected as outliers. However, these works assume a continuous frame as the input, and their application is limited to change detection in video frames. \cite{Gueguen2015} proposed a semi-supervised method to detect damaged areas from pairs of satellite images. In their method, a bag-of-visual-words vector is extracted for hierarchical shape descriptors and a support vector machine classifier is trained on the extracted features. Since their method is based on the carefully chosen feature descriptors specialized for their task, the method lacks generalizability for application in other domains.

Disentangled representation learning is an actively studied field. \cite{Tran2017} proposed a generative adversarial network (GAN) framework to learn disentangled representation for pose and identity of a face using encoder-decoder architecture with auxiliary variables inserted in its latent code. \cite{Odena2016} proposed a GAN model that can generate synthetic images conditioned on category labels. \cite{Siddharth2017} proposed a semi-supervised method to learn disentangled representation by introducing graphical model structure between the encoder and decoder of a standard variational auto-encoder (VAE). A drawback of these methods is that during training, they require explicit labels for the target factor of variation. As for an unsupervised approach, \cite{Chen2016} proposed a method that learns disentangled representation by maximizing mutual information between a small subset of latent codes and a generated image. However, this method cannot control the disentanglement so that the desired factor of variations is represented in a certain latent code. Some works utilize groups of observations as weak supervision. \cite{Kulkarni2015} trains a target latent unit on grouped mini-batches that include only one factor of variation. \cite{Bouchacourt2017} and \cite{Mathieu2016} proposed a method that effectively disentangles intra-class and inter-class variations using groups of images sharing the same class labels. Our work is similar to the three works mentioned above. The difference is that our work assumes weaker conditions; that is, our method only requires pairs of observations and does not require aligned observations or class labels. Recently, multi-view image generation method that only use a paired observation for feature disentanglement is proposed in \cite{chen2017multi}.

Our work is technically inspired by \cite{Bousmalis2016}. The method by \cite{Bousmalis2016} learns common and specific features between two different image domains. The key difference between their work and ours is that, in event detection tasks, the images in a pair come from the ``same'' domain. Since there are no domain biases in the images, we cannot resort to adversarial discriminators when we learn common features. Instead, a distance function in the feature space is used in our work. Since our method is based on a probabilistic latent variable model, rich information of posterior distribution can be used for measuring a distance between features. This is an advantage of using VAE instead of the classical auto-encoders used by \cite{Bousmalis2016}.

\section{Methods}
\label{sect:methods}

\subsection{Overview}
\label{sect:overview}
Figure~\ref{fig:method_rep_learning} shows a schematic of the proposed model. The model consists of two branches of VAEs that share parameters each other. Each VAE extracts two types of feature representations: {\it common} and {\it specific}. They represent different aspects of an input image pair (invariant and variant factors, respectively). In the context of rare event detection, the specific features represent trivial events, and the common features represent image contents that are invariant to trivial events. In order to achieve disentanglement, we introduce a similarity constraint between common features. This constraint promotes common features to lie in a shared latent space of paired images. The key aspect of the common features is that they are invariant to trivial events, which should be helpful to distinguish target events from trivial events. In the successive fine-tuning phase, an event detector is trained on the learned common features using a small number of positive and negative samples.
  
The contents of this section are as follows. In Section~\ref{sect:vae}, we present a brief introduction to VAEs. In Section~\ref{sect:rep_learning}, the proposed method of representation learning is explained in detail. Finally, in Section~\ref{sect:fine-tuning}, the fine-tuning phase of the event detector is explained.
 
\subsection{Variational Auto-encoder}
\label{sect:vae}
A variational auto-encoder \cite{Kingma2014a,Rezende2014} is a kind of deep generative model that defines the joint distribution of an input $\bm{x}\in　\bm{X}$ and a latent variable $\bm{z}\in \bm{Z}$ as $p_{\bm{\theta}}(\bm{x},\bm{z})=p_{\bm{\theta}}　(\bm{x}|\bm{z})p(\bm{z})$. $p(\bm{z})$ is often set to be Gaussian distribution with zero mean and unit variance. The generative distribution $p_{\bm{\theta}}(\bm{x}|\bm{z})$ is modeled by a deep neural network (decoder) with parameters $\bm{\theta}$, and the model parameters are trained by maximizing the marginal likelihood $p_{\bm{\theta}} (\bm{x})=\sum_{\bm{z}} p_{\bm{\theta}} (\bm{x},\bm{z})$. However, in the case that $p_{\bm{\theta}} (\bm{x}|\bm{z})$ is a neural network, the marginal likelihood becomes intractable. Therefore, the following variational lower bound is used instead:
\begin{equation}
\begin{split}
\mathcal{L}_{VAE}=\mathbb{E}_{q_{\bm{\phi}} (\bm{z}|\bm{x})} [\log⁡{p_{\bm{\theta}}(\bm{x}|\bm{z})}]\qquad\qquad\qquad\\
\quad -D_{KL} (q_{\bm{\phi}}(\bm{z}|\bm{x}) \parallel p(\bm{z}))\leq \log{p_{\bm{\theta}}(\bm{x})}
\end{split}
\label{eq:elbo}
\end{equation}
In the above equation, $q_{\bm{\phi}} (\bm{z}|\bm{x})$ is another deep neural network (encoder) that approximates the posterior distribution $p_{\bm{\theta}} (\bm{z}|\bm{x})$. The first term of Eq.~(\ref{eq:elbo}) can be seen as the reconstruction error of a classical auto-encoder, and the second term can be seen as the regularization term. In order to make the lower bound differentiable in terms of the encoder parameters, a technique called reparameterization is used:
\begin{equation}
\bm{z}=\mu_{\bm{\phi}} (\bm{x})+\sigma_{\bm{\phi}} (\bm{x})\odot\bm{\epsilon }\quad{\rm where}\quad \bm{\epsilon}\sim\mathcal{N}(\bm{0},\bm{1})
\label{eq:reparam}
\end{equation}
Here, $\odot$ represents an element-wise product. In this case, the encoder becomes a deep neural network that outputs mean and variance of the posterior distribution.

\subsection{Representation Learning}
\label{sect:rep_learning}
VAE provides an unsupervised method to learn latent representations. Given input $\bm{x}$, the latent representation can be inferred using the encoder distribution $q_{\bm{\phi}} (\bm{z}|\bm{x})$. The objective here is to learn the encoder distribution $q_{\bm{\phi}} (\bm{z}_{c},\bm{z}_{s}|\bm{x})$ in which latent variables are disentangled such that $\bm{z}_c$ and $\bm{z}_s$ respectively represent invariant and variant factors in a given image pair. For this, we build a model with two branches of VAEs that share parameters each other. As shown in Figure~\ref{fig:method_rep_learning}, the input images $\bm{x}_{A},\bm{x}_{B}\in \bm{X}$ are fed into different VAE branches, and the latent variables $\bm{z}_c$ and $\bm{z}_s$ are extracted from each branch. The parameters of VAEs are trained using following loss function.
\begin{equation}
\mathcal{L}=\mathcal{L}_{VAE}^{A}+\mathcal{L}_{VAE}^{B}+\lambda_{1}\mathcal{L}_{sim}+\lambda_{2}\mathcal{L}_{act}
\label{eq:total_loss}
\end{equation}
where $\mathcal{L}_{VAE}^{A},\mathcal{L}_{VAE}^{B}$ are VAE losses for input images $\bm{x}_A$ and $\bm{x}_B$, respectively. $\mathcal{L}_{sim}$ is a similarity loss function that constrains common features to represent invariant factors between paired images. $\mathcal{L}_{act}$ is an activation loss function that encourages activation of common features to avoid a trivial solution. $\lambda_1$ and $\lambda_2$ are the coefficients of similarity and activation loss, respectively. The following explains the details of each type of loss.
\newline\newline
{ \bf Variational auto-encoder loss.} The joint distribution of each VAE branch becomes
\begin{equation}
p_{\bm{\theta}} (\bm{x},\bm{z}^{c},\bm{z}^{s}) = p_{\bm{\theta}} (\bm{x}|\bm{z}^{c},\bm{z}^{s}) p(\bm{z}^{c}) p(\bm{z}^{s})
\label{eq:generative_model}
\end{equation}
The generative distribution $p_{\bm{\theta}} (\bm{x}|\bm{z}^{c},\bm{z}^{s})$ is set as a Gaussian distribution with its mean given by the decoder output. The priors $p(\bm{z}^{c})$ and $p(\bm{z}^{s})$ are both Gaussian distributions with zero mean and unit variance. Then, the inference model becomes
\begin{equation}
q_{\bm{\phi}} (\bm{z}^{c},\bm{z}^{s}|\bm{x}) = q_{\bm{\phi}} (\bm{z}^{c}|\bm{x}) q_{\bm{\phi}} (\bm{z}^{s}|\bm{x})
\label{eq:inference_model}
\end{equation}
The posteriors for $\bm{z}_c$ and $\bm{z}_s$ are set to Gaussian distributions $q_{\bm{\phi}} (\bm{z}^{c}|\bm{x})=\mathcal{N}(\mu_{\bm{\phi}}^{c} (\bm{x}),\sigma_{\bm{\phi}}^{c} (\bm{x}))$ and $q_{\bm{\phi}} (\bm{z}^{s}|\bm{x})=\mathcal{N}(\mu_{\bm{\phi}}^{s} (\bm{x}),\sigma_{\bm{\phi}}^{s} (\bm{x}))$ whose mean and variance are given by the outputs of encoder networks. Then, the loss function of the VAE becomes
\begin{equation}
\begin{split}
\mathcal{L}_{VAE}=\mathbb{E}_{q_{\bm{\phi}} (\bm{x},\bm{z})} [\log{ p_{\bm{\theta}} (\bm{x}|\bm{z})}]\qquad\qquad\qquad\qquad\qquad \\
+D_{KL} (q_{\bm{\phi}} (\bm{z}^{c}|\bm{x}) \parallel (\bm{z}^{c}))+D_{KL} (q_{\bm{\phi}}(\bm{z}^{s}|\bm{x}) \parallel (\bm{z}^{s}))
\end{split}
\label{eq:loss_vae}
\end{equation}
\newline\newline
{ \bf Similarity loss.} In order to make common features encode invariant factors in an input image pair, we introduce the following similarity loss between the pair of common features extracted from $\bm{x}_A$ and $\bm{x}_B$:
\begin{equation}
\mathcal{L}_{sim}=D(q_{\bm{\phi}} (\bm{z}^{c}|\bm{x}_{A}) \parallel q_{\bm{\phi}} (\bm{z}^{c}|\bm{x}_{B}))
\label{eq:loss_sim}
\end{equation}
where $D$ defines a distance between latent variables $\bm{z}_c$ and $\bm{z}_s$. There are various types of similarity metric that can be used for $D$. One possible choice is a kind of geometric distance ({\it e.g.,} L2 or L1 norm) between the mean vectors $\mu(\bm{x})$ of the posterior distribution. However, they produce inaccurate distance measures because they do not take the shape of the posterior distribution into account. Even if the distance between the two mean vectors is large, they are semantically near in the case that the posterior has flat distribution. On the contrary, even if the distance between the two mean vectors is small, they are semantically far away in the case that the posterior has sharp distribution. Therefore, in this work, we also take the standard deviation $\sigma(\bm{x})$ into consideration and use the following distance function called Modified-L2:
\begin{equation}
\mathcal{L}_{sim}=\frac{1} {M} \sum_{i}^{M} \frac{(\mu_{A,i}^{c}-\mu_{B,i}^{c} )^2} {\sigma_{A,i}^{c}\sigma_{B,i}^{c}}
\label{eq:loss_sim_ml2}
\end{equation}
Here, $\mu_{*,i}^c$ and $\sigma_{*,i}^c$ represent the $i$-th element of the mean and standard deviation vector, and $M$ is the dimension of the vector. Alternatively, we could use a distance between distributions such as f-divergences. For more detail and results of the comparison, see Section~\ref{sect:ablation}.
\newline\newline
{ \bf Activation loss.} One problem with the similarity constraints is that there exists a trivial solution. The model can completely satisfy the constraints by setting the mean vectors of common features to all zeros. In this case, all the information in the input are encoded by specific features and common features do not represent any information. To avoid this, we introduce another loss to encourage activation of common features:
\begin{equation}
\mathcal{L}_{act}=\mathcal{L}_{sparsity}+\mathcal{L}_{invmax}
\label{eq:loss_act}
\end{equation}
Activation loss consists of two parts: sparsity loss and invmax loss:
\begin{equation}
\mathcal{L}_{sparsity}=\sum_{i=1}^{d} (s\log⁡{m_i}+(1-s)\log⁡{(1-m_i)})
\end{equation}
\begin{equation}
\mathcal{L}_{invmax}=\frac{1}{B} \sum_{k=1}^{B} (\max_i{|\mu_{i}^{k}⁡|})^{-1} 
\end{equation}
Here, $\mu_{i}^{k}$ is the $i$-th element of the mean vector for the $k$-th input sample, and $m_i$ is the average of the absolute values of $\mu_{i}^{k}$ in the mini-batch, {\it i.e.,} if the mini-batch size is $B$, $m_i=\sum_{k=1}^{B} |\mu_{i}^{k}|$. In this activation loss, $\mathcal{L}_{sparsity}$ constrains the absolute mean of activation values to be close to a hyperparameter $s$, and $\mathcal{L}_{invmax}$ encourages there to be at least one unit of mean vectors activated for each sample. 

\subsection{Fine-tuning}
\label{sect:fine-tuning}
Now we have acquired the encoder to extract disentangled representation for variant and invariant factors of a given image pair. As the next step, we now introduce the fine-tuning phase. In this phase, we build a event detector network $C_{\bm{\psi}}$ on common features for paired input $\bm{x}_A$, $\bm{x}_B$. Specifically, the mean vector $\bm{\mu}_{A}^{c}$,$\bm{\mu}_{B}^{c}$ of common features is used as input to the classifier.
\begin{equation}
\bm{y}=C_{\bm{\psi}} (\bm{\mu}^{c}) \quad{\rm where}\quad \bm{\mu}^{c}=[\bm{\mu}_{A}^{c},\bm{\mu}_{B}^{c}]
\label{eq:classifier_model}
\end{equation}
Here, [*,*] represents a concatenation of two vectors. Cross-entropy loss is used to train the classifier on ground truth label $\bm{t}$.
\begin{equation}
\mathcal{L}_{fine}=\bm{t}\log{\bm{y}}+(1-\bm{t})\log{(1-\bm{y})}
\label{eq:loss_fine}
\end{equation}
In the fine-tuning phase, classifier parameters $\bm{\psi}$ and encoder parameters $\bm{\phi}$ are jointly trained. Because common features are trained to be invariant to trivial events in the representation learning phase, we can train the event detector, which is robust to trivial events. During the fine-tuning, negative samples are randomly under-sampled to have the same number of samples as positives.

\section{Experiments}
\label{sect:experiments}
In Section~\ref{sect:results}, we conducted the qualitative evaluations on four change detection datasets: Augmented MNIST, ABCD, PCD, and WDC dataset. First, we verified our method on Augmented MNIST dataset, and then evaluated our method on in-the-wild datasets (ABCD, PCD, and WDC dataset). While all the datasets originally contained many positive samples, we limited the available positive samples to simulate a class-imbalance scenario. The numbers of positive and negative samples used in the experiments are listed in Table~\ref{tb:dataset}. In Section~\ref{sect:vis_analysis}, we conducted the quantitative evaluation by visualizing learned features on Augmented MNIST dataset. Finally, in Section~\ref{sect:ablation}, we investigated several design choices of our model on Augmented MNIST and ABCD dataset.

\subsection{Datasets}
\label{sect:datasets}

\noindent\newline
{ \bf Augmented MNIST.} To validate the proposed model, we set a problem of detecting a change of digit from a pair of samples in MNIST. An input image pair was labeled as positive if the digits in each image were different and labeled as negative if they were same. For source images, we used three variants of MNIST \cite{A-MNIST}: MNIST with rotation (MNIST-R), background clutter (MNIST-B), and both (MNIST-R-B). 

\noindent\newline
{ \bf ABCD dataset.} The ABCD dataset \cite{Fujita2017} is a dataset for detecting changes in buildings from a pair of aerial images. The aerial images have 40 cm spatial resolution and are taken before and after a tsunami disaster. The task is to classify whether the target buildings were washed away by the tsunami or not. Training and test patches were resized and cropped in advance such that the target buildings were in the center ({\it i.e.,} we used ``resized'' patches as used in \cite{Fujita2017}).

\noindent\newline
{ \bf PCD dataset.} The PCD dataset \cite{Sakurada2015} is a dataset for detecting scene changes from a pair of street view panorama images. For each pair, pixel-wise change masks are provided as ground truth. In this work, we solved the change mask estimation problem by conducting patch-based classification. First, input patch pairs of size $112\times112$ were cropped from original images, then they were labeled as positive if the center area of size $14\times14$ was purely changed pixels and labeled as negative if the center area was purely unchanged pixels. In the testing phase, we cropped the patch pairs in a sliding manner, and overlaid the classifier outputs to create a heatmap of change probabilities. The heatmap was then binarized using a threshold of 0.5, which results in change mask estimation.

\noindent\newline
{ \bf WDC dataset.} In order to evaluate our method on a more large scale dataset, we prepared the new change detection dataset. This dataset is for detecting newly constructed or destructed buildings from a pair of aerial images. For the dataset, we used the aerial images and the building footprints of Washington D.C. area. The aerial images are taken at different years (1995, 1999, 2002, 2005, 2008, 2010, 2013 and 2015). The images have 16 cm resolution, and covers over 200 $km^2$ for each year. We automatically annotate changes in buildings by comparing building footprints produced at different years. All the source data are acquired from open data repository hosted by the Government of District of Columbia \cite{WDC}. For more detailed information about the dataset, please refer to the supplementary material.


\subsection{Experimental setup}
\label{sect:setups}

\begin{table}[tb]
\caption{Number of positive and negative samples used in each dataset. All the negative samples were used for representation learning. In fine-tuning, both the negative and positive samples were used.}
\label{tb:dataset}
\centering
\scalebox{0.75}{
\centering
\begin{tabular}{c|cc|cc}
\hline
                & \multicolumn{2}{c|}{Training}   & \multicolumn{2}{c}{Testing} \\
                & \#negatives & \#positives       & \#negatives  & \#positives  \\ \hline
Aug. MNIST & 100,000     & 50 / 500 / 32,000 & 50,000       & 50,000       \\
ABCD            & 3374        & 5 / 50 / 3378   & 847          & 845          \\
PCD             & 56718       & 50               & -            & -            \\
WDC             & 250,000       & 50 / 500    & 1934   & 1934    \\ \hline
\end{tabular}
}
\end{table}

\begin{table}[tb]
\caption{Comparison to the anomaly detection methods on Augmented MNIST dataset. For all models, only negative samples were used during training.}
\label{tb:result_mnist_unsupervised}
\centering
\scalebox{0.9}{
\centering
\begin{tabular}{c|ccc}
\hline
                   & MNIST-R        & MNIST-B        & MNIST-R-B      \\ \hline
AE-rec \cite{xia2015learning}            & 54.27          & 54.48          & 51.36          \\
VAE-rec \cite{an2015variational}           & 57.24          & 53.27          & 50.7           \\
CAE-l2 \cite{aytekin2018clustering}            & 55.14          & 55.74          & 50.29          \\
MLVAE \cite{Bouchacourt2017}             & 60.72          & 59.70          & 52.75          \\
Mathieu et al. \cite{Mathieu2016}    & 58.34          & 60.31          & 52.16          \\
VAE w/o sim.       & 54.95          & 56.44          & 52.02          \\
VAE w/ sim. (ours) & \textbf{71.66} & \textbf{82.55} & \textbf{62.23} \\ \hline
\end{tabular}

}
\end{table}

\begin{table*}[tb]
\caption{Change detection accuracies on Augmented MNIST dataset. The number of positive samples were varied from 50 to 32,000. Each result is given in terms of the mean and standard deviation obtained by 10 training runs using different training subsets.}
\label{tb:result_mnist}
\centering
\scalebox{0.9}{
\centering
\begin{tabular}{cc|cccccc}
\hline
          & \#Labels & Under samp.  & Over samp.   & MLVAE \cite{Bouchacourt2017}       & Mathieu et al. \cite{Mathieu2016} & VAE w/o sim. & VAE w/ sim. (ours) \\ \hline
MNIST-R   & 50      & 50.63(±0.31) & 50.47(±0.44) & 57.22(±1.39) & 61.09(±1.20)   & 51.55(±0.43) & \textbf{79.65(±4.42)}       \\
          & 500     & 60.05(±3.10) & 61.84(±1.37) & 79.15(±0.90) & 77.78(±0.74)   & 64.74(±1.31) & \textbf{89.73(±0.56)}       \\
          & 32000    & 94.82(±0.21) & 95.49(±0.15) & 95.68(±0.17) & 95.85(±0.23)   & 95.76(±0.09) & \textbf{95.94(±0.15)}       \\ \hline
MNIST-B   & 50      & 50.69(±0.61) & 50.38(±0.16) & 59.33(±2.25) & 58.79(±2.66)   & 52.67(±1.44) & \textbf{82.16(±0.37)}       \\
          & 500     & 52.04(±1.52) & 52.27(±2.80) & 72.26(±0.96) & 75.16(±1.09)   & 73.56(±2.24) & \textbf{84.69(±0.42)}       \\
          & 32000    & 94.92(±0.21) & 93.28(±0.15) & 95.67(±0.10) & 94.47(±0.29)   & \textbf{96.25(±0.06)} & 96.05(±0.13)       \\ \hline
MNIST-R-B & 50      & 50.30(±0.11) & 50.37(±0.08) & 51.61(±0.67) & 51.19(±0.51)   & 50.32(±0.28) & \textbf{60.58(±1.60)}       \\
          & 500     & 50.35(±0.12) & 50.47(±0.19) & 56.21(±0.27) & 53.10(±0.93)   & 52.39(±0.49) & \textbf{62.68(±0.46)}       \\
          & 32000    & 79.04(±0.25) & 75.94(±0.80) & 78.73(±0.26) & 78.55(±1.17)   & 80.92(±0.41) & \textbf{81.54(±0.57)}       \\ \hline
\end{tabular}
}
\vspace{-0.5mm}
\end{table*}
\begin{table*}[tb]
\caption{Change detection accuracies on the ABCD, WDC and PCD dataset. On the column of ABCD and WDC dataset, accuracies are presented for different numbers of positive samples. On PCD dataset, the performance is reported for three evaluation metrics (Accuracy, mIoU, and IoU for positive class). The number of positive samples used for PCD dataset is 50. Each result is given in terms of the mean and standard deviation obtained by 10 training runs using different training subsets.}
\label{tb:result_abcd_pcd_wdc}
\centering
\scalebox{0.8}{
\centering
\begin{tabular}{c|ccc|cc|ccc}
\hline
                    & \multicolumn{3}{c|}{ABCD}                                             & \multicolumn{2}{c|}{WDC}                      & \multicolumn{3}{c}{PCD}                                              \\
                    & \#Labels 5           & 50                   & All                   & \#Labels 50          & 500                  & Acc.                  & mIoU                  & IoU                   \\ \hline
Under samp.         & 61.14(±11.61)         & 64.05(±17.16)         & 95.24(±0.20)          & 53.12(±4.56)          & 51.72(±3.03)          & 73.28(±3.10)          & 56.27(±3.32)          & 47.95(±2.20)          \\
Over samp.          & 60.88(±13.58)         & 54.05(±11.78)         & 92.91(±0.39)          & 52.02(±3.37)          & 52.09(±4.80)          & \textbf{80.52(±3.48)} & 60.88(±3.68)          & 44.92(±3.49)          \\
Transfer            & 77.39(±7.30)          & 88.17(±0.75)          & \textbf{96.03(±0.19)} & 61.32(±1.73)          & 71.07(±3.04)          & 75.59(±2.58)          & 58.74(±2.77)          & 49.60(±2.18)          \\
MLVAE \cite{Bouchacourt2017}              & 65.36(±5.19)          & 86.31(±1.80)          & 95.33(±0.19)          & \textbf{63.58(±1.59)} & 74.70(±0.77)          & 76.88(±1.22)          & 60.13(±1.50)          & 50.55(±1.75)          \\
Mathieu et al. \cite{Mathieu2016}     & 64.73(±5.41)          & 77.66(±2.11)          & 91.79(±0.21)          & 60.54(±2.80)          & 71.55(±0.69)          & 73.71(±3.55)          & 56.63(±3.59)          & 48.02(±2.13)          \\
VLAE w/o sim.       & 67.32(±6.51)          & 86.69(±1.79)          & 95.18(±0.14)          & 59.41(±1.68)          & 74.17(±1.05)          & 77.22(±1.75)          & 60.49(±2.27)          & 50.73(±2.70)          \\
VLAE w/ sim. (ours) & \textbf{78.52(±5.01)} & \textbf{89.70(±0.77)} & 95.60(±0.14)          & 63.25(±0.86)          & \textbf{75.70(±0.66)} & 78.20(±1.96)          & \textbf{61.66(±2.23)} & \textbf{51.77(±1.84)} \\ \hline
\end{tabular}
}
\vspace{0mm}
\end{table*}

\begin{figure*}[tb]
\vspace{-1mm}
\centering
\includegraphics[width=17cm]{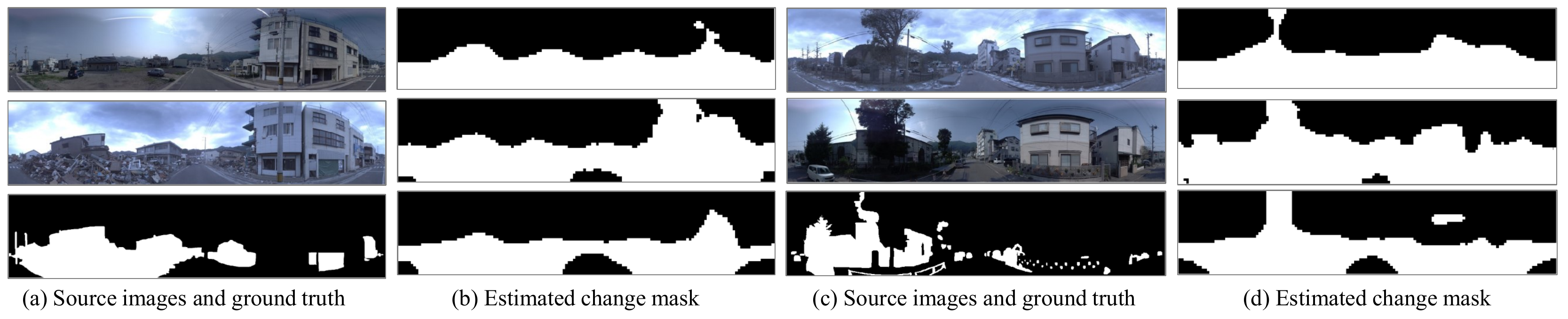}
\caption{Examples of mask estimation results on the PCD dataset. From top to bottom, the figures in the columns {\it b} and {\it d} shows the result of ``Under samp.'', ``Transfer'', and ``VLAE w/ sim. (ours)'', respectively.}
\label{fig:mask_pcd}
\end{figure*}

\noindent\newline
{\bf Baselines.} For comparison, we built several baseline models for handling the class-imbalance problem. (1) {\it Random under/over-sampling}: a straightforward approach for class-imbalance problem is under-sampling of major class instances or over-sampling of minor class instances. For the approach, we trained a variant of siamese CNN (the state-of-the-art architecture for image comparison tasks) with the sampling schemes. (2) {\it Transfer learning}: transfer learning is considered to be effective when the number of available labels are limited. We transferred weights from the ImageNet pre-trained models, and fine-tuned it with under-sampling scheme. (3) {\it Disentangled representation learning methods}: For comparison with the state-of-the-art representation learning models, we replaced our model with the models proposed in \cite{Bouchacourt2017,Mathieu2016}. In the original formulation of \cite{Mathieu2016}, the discriminator requires class labels as its additional input. In our experiments, we used image pair instead ({\it i.e.,} discriminate real-generated and real-real pairs) since we have no access to the class labels. (4) {\it Anomaly detection methods}: we also built several anomaly detection methods from \cite{an2015variational,aytekin2018clustering,xia2015learning}. To apply the methods, each image pair is concatenated and regarded as single data point. The models are trained using only negative ({i.e. normal}) data, and rare events are detected as outliers.

\noindent\newline
{\bf Model architecture for representation learning.} We built two architectures: one for Augmented MNIST dataset and another for the ABCD and PCD datasets. For Augmented MNIST dataset, the encoder had a simple architecture of ``C-P-C-P-C-H'', where C, P, and H represent convolution, max-pooling, and hidden layer, respectively. Here, the hidden layer consists of four branches of convolutional layers, which extract mean and log-variance of specific and common features. For the ABCD and PCD dataset, in order to model complex real-world scenes, we used a hierarchical latent variable model proposed in \cite{Zhao2017}, where a particular image is modeled by a combination of multiple latent variables with different levels of abstraction. Specifically, we used a model with 5 hidden layers in the experiments. Because target events are often related to high-level image contents, the common features were extracted only on the top two hidden layers. For both architectures above, the decoder part was set to be symmetric to its encoder. For the detailed architecture and the hyper-parameter settings, please refer to the supplementary materials.

\noindent\newline
{\bf Model architecture for fine-tuning.} 
In the fine-tuning phase, we attached an event detector consisting of three fully-connected layers. The dimensions of the layers were 100-100-2 for Augmented MNIST dataset and 2048-2048-2 for the ABCD and PCD datasets. During fine-tuning, the learning rate of the pre-trained encoder part was down-weighted by a factor of 10.

\begin{figure*}[tb]
\begin{tabular}{c}

\begin{minipage}{0.97\hsize}
\centering
\includegraphics[width=14.5cm]{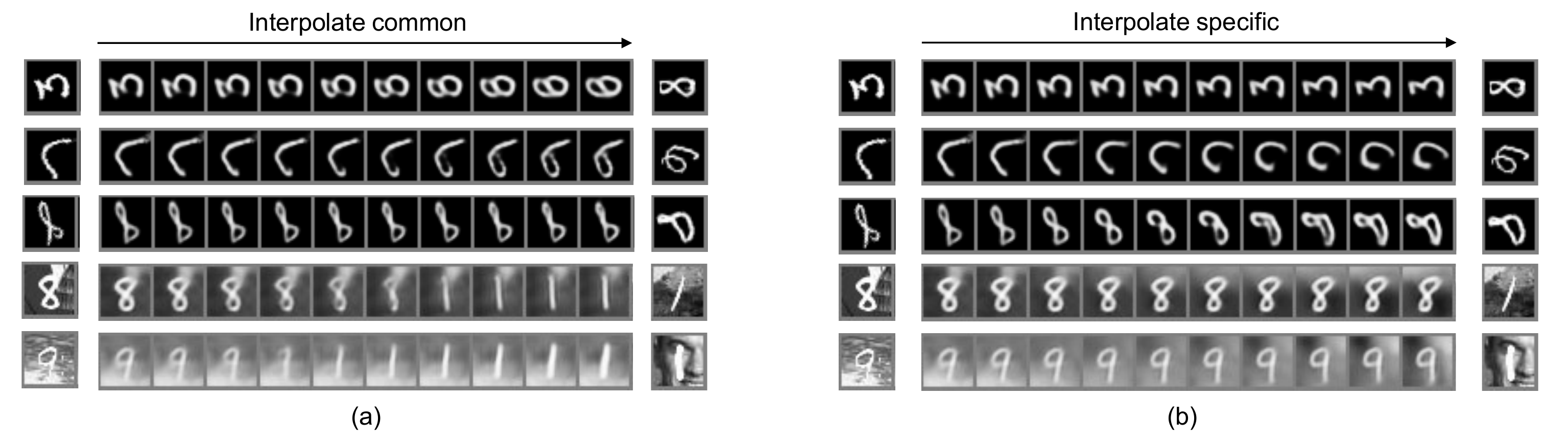}
\caption{Results of feature interpolation analysis. \textbf{(a)} Interpolation of common features. \textbf{(b)} Interpolation of specific features.}
\label{fig:vis_interp}
\end{minipage} \\

\begin{minipage}{0.97\hsize}
\vspace{2mm}
\end{minipage} \\

\begin{minipage}{0.97\hsize}
\centering
\includegraphics[width=16.5cm]{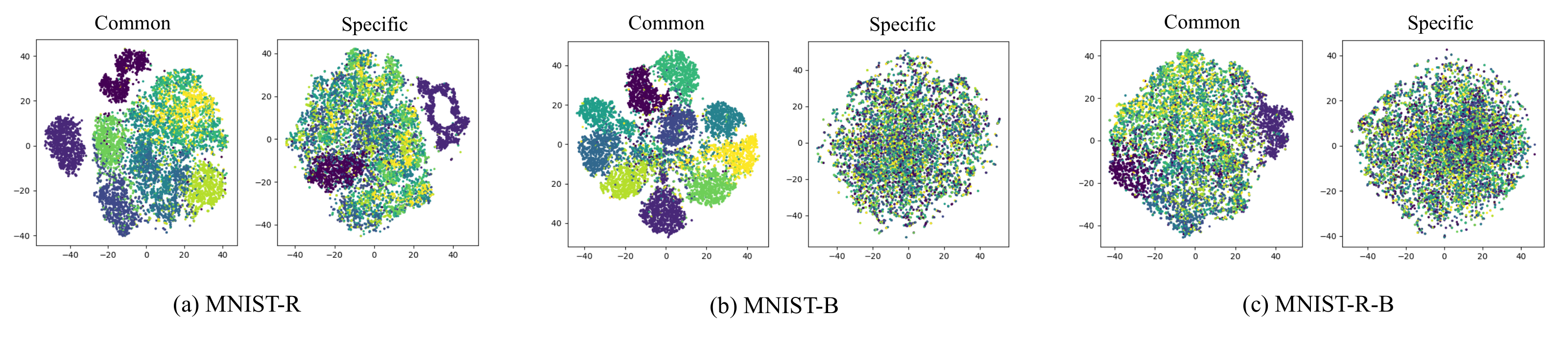}
\caption{Results of t-SNE visualization for common and specific features. The color of each plot represents digit classes.}
\label{fig:vis_tsne}
\end{minipage}

\end{tabular}
\vspace{0mm}
\end{figure*}

\subsection{Quantitative results}
\label{sect:results}
Table~\ref{tb:result_mnist} shows the results for Augmented MNIST dataset. When labels were scarce, the proposed method outperformed the other models by a large margin. By comparing the models with and without similarity loss (``VAE w/o sim.'' and ``VAE w/ sim.''), we can conclude that the proposed similarity loss is essential to learn better representations for the change detection task. The performance improvement is especially remarkable with 50 labels, where the proposed model improved by approximately 20-30\% compared to the baselines.

In Table~\ref{tb:result_mnist_unsupervised}, we also compare our method to several anomaly detection methods. In this case, we did not train the event detector. Instead, we detected change events by applying k-means clustering to the distance between common features. In the table, the proposed method outperforms the other models.

Table~\ref{tb:result_abcd_pcd_wdc} shows the results for the ABCD, WDC and PCD datasets, respectively. Also, for these in-the-wild datasets, the proposed method outperformed the other baselines. Figure~\ref{fig:mask_pcd} compares the estimated change mask for baseline models and that of the proposed model. We see that the baseline models are sensitive to illumination changes or registration errors in roads or buildings. Clearly, they suffer from false alarms created by trivial events. On the other hand, much of the false alarms were successfully suppressed in the output of the proposed model.

\subsection{Visualization of Latent Variables}
\label{sect:vis_analysis}
In this subsection, we investigate what is learned in common features and specific features by conducting several analyses. {\it Interpolation}: to visually evaluate the disentanglement, we generated a sequence of images by linearly interpolating image representations between pairs of images. To independently investigate the learned semantics of common and specific features, the features were interpolated one at a time while fixing the others. Figure~\ref{fig:vis_interp} shows the result of visualization on Augmented MNIST dataset. When common features were interpolated between different digits, the digit classes in the generated sequences gradually changed accordingly, while the other factors ({\it i.e.,} rotation, styles, and background) were unchanged. On the other hand, when specific features are interpolated, rotation angles or background patterns are changed accordingly, while the digit classes remained the same. The result shows that the common features extract information about digit classes, but they are invariant to the variation observed in the same digit pairs. {\it 2D visualization:} we visualized the learned features by t-SNE \cite{Maaten2008}. Figure~\ref{fig:vis_tsne} shows the visualization results for common and specific features. In this figure, the same color plots correspond to the same digits. We see that the common features are more informative about digit classes compared to the specific features.

We also conducted the above visualization for the rest of the datasets. However, for the real-world complicated scenes, it was difficult to achieve clear disentanglement. Specifically, we observed that the activations of the units in the common features are degenerated to a certain value.

\subsection{Ablation Study}
\label{sect:ablation}
\begin{figure}[tb]
\centering
\hspace{-2mm}\includegraphics[width=8cm]{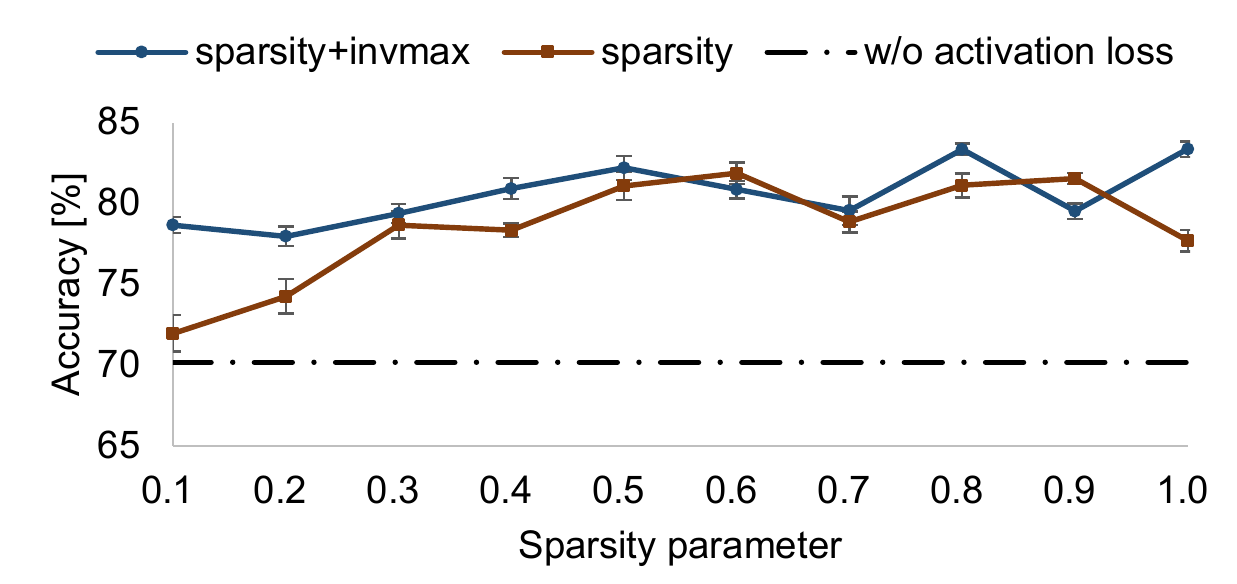}
\caption{Analysis of the effect of activation loss. Results for different sparsity parameter values are shown with red plots (without invmax-loss) and blue plots (with invmax-loss). The error bar shows the standard deviations of the accuracies for 10 runs with different training subsets}
\label{fig:ablation_act}
\end{figure}

\begin{figure}[tb]
\centering
\vspace{-1mm}
\hspace{-2mm}\includegraphics[width=7cm]{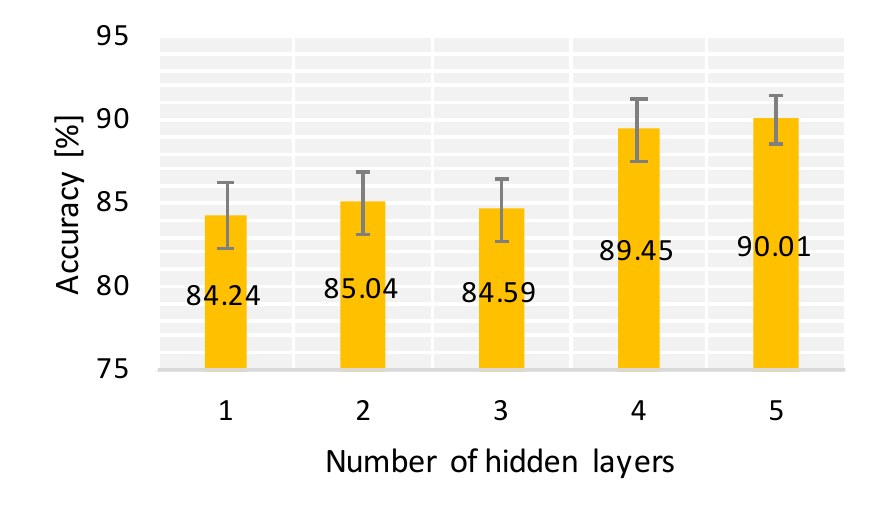}
\vspace{-3mm}
\caption{Sensitivity analysis of the number of hidden layers. The error bar shows the standard deviations of the accuracies for 10 runs with different training subsets.}
\label{tb:ablation_number_of_hidden}
\end{figure}

{ \bf Effect of activation loss.} To investigate the effect of activation loss (Eq.~(\ref{eq:loss_act})), we conducted sensitivity analysis on the sparsity parameter and existence of $\mathcal{L}_{invmax}$. Figure~\ref{fig:ablation_act} shows the results for MNIST-R. From these results, we can draw several conclusions. First, a sparsity parameter of approximately 0.5 seems to be suitable because the performance becomes unstable in terms of the choice of parameter value at larger than 0.5. Secondly, the use of $\mathcal{L}_{invmax}$ boosts performance. Lastly and most importantly, regardless of parameter choices, the presence of activation loss improves performance.
\newline\newline
{ \bf Choice of the distance function for similarity loss.} Here, we investigate several choices of distance function in Eq.~(\ref{eq:loss_sim}). Table~\ref{tb:ablation_sim_metric} compares six types of distance function evaluated on the MNIST-R and ABCD datasets. We found that both Modified-L2 and Jeffrey’s divergence are suitable choices. This result supports our intuition that we should consider not only the mean vector of the latent distribution but also the shape of the distribution.
\newline\newline
{ \bf Importance of hierarchical latent variables.} Here, we investigate the effect of using hierarchical latent variable models for representation learning. In this analysis, the hidden layers were eliminated one by one from the proposed model with 5 hidden layers. The elimination was conducted in order of lowest layer to highest. Figure~\ref{tb:ablation_number_of_hidden} shows the results on the ABCD dataset. In the figure, the models with 4 and 5 hidden layers perform better. This result shows the importance of extracting hierarchical latent variables in rare event detection tasks. A pair of scenes includes various types of events, which range from low-level to high-level. By utilizing hierarchical latent variables, the events of each level can be represented by each level of latent variables. As a result, detection models can easily distinguish high-level events from low-level ones. The other merit is that we can choose which level of features to use according to target events.

\section{Conclusion}
\label{sect:conclustion}
We proposed a novel representation learning method to overcome the class-imbalance problem in rare event detection tasks. 
The proposed network learns the two separated features related to image contents and other nuisance factors from only low-cost negative samples by introducing a similarity constraint between the image contents.
The learned features are utilized in the subsequent fine-tuning phase, where rare event detectors are learned robustly. 
The effectiveness of the proposed method was verified by the quantitative evaluations on the four change detection datasets. For the the evaluations, we created a large-scale change detection dataset using publicly available data repository. In addition, the qualitative analysis on Augmented MNIST showed that the model successfully learns the desired disentanglement. 

The disentanglement of the proposed method is still insufficient for complicated scenes in the real world, due to degenerated solution observed in the common features. The performance of our method will be greatly improved with the clearer feature disentanglement. A possible next step to achieve this is to avoid degenerated solution by introducing adversarial training as used in \cite{Mathieu2016}, or maximizing mutual information between the common feature and input images \cite{Chen2016}.
Also, in the future, we intend to apply the learned invariant features to various types of event detection tasks including change mask estimation and change localization.

\begin{table}
\centering
\caption{Comparison of the choices of distance function used in similarity loss.}
\label{tb:ablation_sim_metric}
\scalebox{0.9}{
\begin{tabular}{c|cc}
\hline
                     & MNIST-R      & ABCD         \\ \hline
L2                   & 82.16(±0.64) & 90.13(±1.31) \\
L1                   & 79.14(±0.90) & 89.01(±1.44) \\
Cosine               & 60.94(±0.97) & 89.63(±1.51) \\
MMD                  & 62.30(±0.50) & 89.69(±1.35) \\
Jeffrey's Divergence & 86.90(±0.32) & 89.85(±0.98) \\
Modified L2          & 89.73(±0.56) & 89.70(±0.77) \\ \hline
\end{tabular}
}
\end{table}

\section*{Acknowledgement}
This paper is based on results obtained from a project commissioned by the New Energy and Industrial Technology Development Organization (NEDO).

{\small
\bibliographystyle{ieee}
\bibliography{main}

\begin{thebibliography}{10}\itemsep=-1pt

\bibitem{an2015variational}
J.~An and S.~Cho.
\newblock Variational autoencoder based anomaly detection using reconstruction
  probability.
\newblock {\em SNU Data Mining Center, Tech. Rep.}, 2015.

\bibitem{aytekin2018clustering}
C.~Aytekin, X.~Ni, F.~Cricri, and E.~Aksu.
\newblock Clustering and unsupervised anomaly detection with l2 normalized deep
  auto-encoder representations.
\newblock {\em arXiv preprint arXiv:1802.00187}, 2018.

\bibitem{babenko2014neural}
A.~Babenko, A.~Slesarev, A.~Chigorin, and V.~Lempitsky.
\newblock Neural codes for image retrieval.
\newblock In {\em ECCV}, pages 584--599. Springer, 2014.

\bibitem{Barnich2011}
O.~Barnich and M.~{Van Droogenbroeck}.
\newblock {ViBe: A universal background subtraction algorithm for video
  sequences}.
\newblock {\em IEEE Transactions on Image Processing}, 20(6):1709--1724, 2011.

\bibitem{Bouchacourt2017}
D.~Bouchacourt, R.~Tomioka, and S.~Nowozin.
\newblock {Multi-Level Variational Autoencoder: Learning Disentangled
  Representations from Grouped Observations}.
\newblock {\em arXiv preprint arXiv:1705.08841}, 2017.

\bibitem{Bousmalis2016}
K.~Bousmalis, G.~Trigeorgis, N.~Silberman, D.~Krishnan, and D.~Erhan.
\newblock {Domain Separation Networks}.
\newblock {\em NIPS}, 2016.

\bibitem{Bowman2016}
S.~R. Bowman, L.~Vilnis, O.~Vinyals, A.~M. Dai, R.~Jozefowicz, and S.~Bengio.
\newblock {Generating Sentences from a Continuous Space}.
\newblock {\em CONLL}, 2016.

\bibitem{chen2017multi}
M.~Chen, L.~Denoyer, and T.~Artieres.
\newblock Multi-view data generation without view supervision.
\newblock {\em ICLR}, 2018.

\bibitem{Chen2016}
X.~Chen, Y.~Duan, R.~Houthooft, J.~Schulman, I.~Sutskever, and P.~Abbeel.
\newblock {InfoGAN: Interpretable Representation Learning by Information
  Maximizing Generative Adversarial Nets}.
\newblock {\em NIPS}, 2016.

\bibitem{chen2015deep}
Z.~Chen, X.~Sun, L.~Wang, Y.~Yu, and C.~Huang.
\newblock {A deep visual correspondence embedding model for stereo matching
  costs}.
\newblock In {\em ICCV}, 2015.

\bibitem{Fujita2017}
A.~Fujita, K.~Sakurada, and T.~Imaizumi.
\newblock {Damage Detection from Aerial Images via Convolutional Neural
  Networks}.
\newblock {\em MVA}, 2017.

\bibitem{fujita2017damage}
A.~Fujita, K.~Sakurada, T.~Imaizumi, R.~Ito, S.~Hikosaka, and R.~Nakamura.
\newblock {Damage Detection from Aerial Umages via Convolutional Neural
  Networks}.
\newblock In {\em MVA}, 2017.

\bibitem{Gueguen2015}
L.~Gueguen and G.~Street.
\newblock {Large-Scale Damage Detection Using Satellite Imagery}.
\newblock {\em CVPR}, 2015.

\bibitem{Khan2017}
S.~Khan, X.~He, F.~Porikli, M.~Bennamoun, F.~Sohel, and R.~Togneri.
\newblock {Learning deep structured network for weakly supervised change
  detection}.
\newblock {\em IJCAI}, 2017.

\bibitem{khan2017learning}
S.~H. Khan, X.~He, F.~Porikli, M.~Bennamoun, F.~Sohel, and R.~Togneri.
\newblock {Learning Deep Structured Network for Weakly Supervised Change
  Detection}.
\newblock In {\em IJCAI}, 2017.

\bibitem{Kingma2015}
D.~Kingma and J.~Ba.
\newblock {Adam: A method for stochastic optimization}.
\newblock {\em ICLR}, 2015.

\bibitem{Kingma2014a}
D.~P. Kingma and M.~Welling.
\newblock {Auto-Encoding Variational Bayes}.
\newblock {\em ICLR}, 2014.

\bibitem{Kulkarni2015}
T.~Kulkarni, W.~Whitney, P.~Kohli, and J.~Tenenbaum.
\newblock {Deep Convolutional Inverse Graphics Network}.
\newblock {\em NIPS}, 2015.

\bibitem{Mathieu2016}
M.~Mathieu, J.~Zhao, P.~Sprechmann, A.~Ramesh, and Y.~LeCun.
\newblock {Disentangling factors of variation in deep representations using
  adversarial training}.
\newblock {\em NIPS}, 2016.

\bibitem{Odena2016}
A.~Odena, C.~Olah, and J.~Shlens.
\newblock {Conditional Image Synthesis With Auxiliary Classifier GANs}.
\newblock {\em arXiv preprint arXiv:1610.09585}, 2016.

\bibitem{Rezende2014}
D.~J. Rezende, S.~Mohamed, and D.~Wierstra.
\newblock Stochastic backpropagation and approximate inference in deep
  generative models.
\newblock {\em arXiv preprint arXiv:1401.4082}, 2014.

\bibitem{sakurada2015change}
K.~Sakurada and T.~Okatani.
\newblock {Change Detection from a Street Image Pair using CNN Features and
  Superpixel Segmentation}.
\newblock In {\em BMVC}, 2015.

\bibitem{Sakurada2015}
K.~Sakurada and T.~Okatani.
\newblock {Change Detection from a Street Image Pair using CNN Features and
  Superpixel Segmentation}.
\newblock {\em BMVC}, 2015.

\bibitem{Siddharth2017}
N.~Siddharth, B.~Paige, J.-W. van~de Meent, A.~Desmaison, N.~D. Goodman,
  P.~Kohli, F.~Wood, and P.~H.~S. Torr.
\newblock {Learning Disentangled Representations with Semi-Supervised Deep
  Generative Models}.
\newblock {\em NIPS}, 2017.

\bibitem{simo2015discriminative}
E.~Simo-Serra, E.~Trulls, L.~Ferraz, I.~Kokkinos, P.~Fua, and F.~Moreno-Noguer.
\newblock {Discriminative learning of deep convolutional feature point
  descriptors}.
\newblock In {\em ICCV}, 2015.

\bibitem{Simonyan2014}
K.~Simonyan and A.~Zisserman.
\newblock {Very deep convolutional networks for large-scale image recognition}.
\newblock {\em CoRR}, 2014.

\bibitem{Stauffer1999}
C.~Stauffer and W.~E.~L. Grimson.
\newblock {Adaptive background mixture models for real-time tracking}.
\newblock {\em CVPR}, 1999.

\bibitem{BMVC2015_127}
S.~Stent, R.~Gherardi, B.~Stenger, and R.~Cipolla.
\newblock {Detecting Change for Multi-View, Long-Term Surface Inspection}.
\newblock In {\em BMVC}, 2015.

\bibitem{NIPS2014_5416}
Y.~Sun, Y.~Chen, X.~Wang, and X.~Tang.
\newblock Deep learning face representation by joint
  identification-verification.
\newblock In {\em NIPS}, pages 1988--1996. Curran Associates, Inc., 2014.

\bibitem{Tran2017}
L.~Tran, X.~Yin, and X.~Liu.
\newblock {Disentangled Representation Learning GAN for Pose-Invariant Face
  Recognition}.
\newblock {\em CVPR}, 2017.

\bibitem{WDC}
\url{http://opendata.dc.gov/pages/dc-from-above}.
\newblock {DC GIS program}.

\bibitem{A-MNIST}
\url{http://www.iro.umontreal.ca/~lisa/twiki/bin/view.cgi/Public/MnistVariations}.
\newblock {Augmented MNIST}.

\bibitem{Maaten2008}
L.~van~der Maaten and G.~Hinton.
\newblock {Visualizing Data using t-SNE}.
\newblock {\em Journal of Machine Learning Research}, 9:2579--2605, 2008.

\bibitem{wen2016discriminative}
Y.~Wen, K.~Zhang, Z.~Li, and Y.~Qiao.
\newblock A discriminative feature learning approach for deep face recognition.
\newblock In {\em ECCV}, pages 499--515. Springer, 2016.

\bibitem{xia2015learning}
Y.~Xia, X.~Cao, F.~Wen, G.~Hua, and J.~Sun.
\newblock Learning discriminative reconstructions for unsupervised outlier
  removal.
\newblock In {\em ICCV}, 2015.

\bibitem{zagoruyko2015learning}
S.~Zagoruyko and N.~Komodakis.
\newblock {Learning to compare image patches via convolutional neural
  networks}.
\newblock In {\em CVPR}, 2015.

\bibitem{zbontar2016stereo}
J.~Zbontar and Y.~LeCun.
\newblock {Stereo matching by training a convolutional neural network to
  compare image patches}.
\newblock {\em JMLR}, 17(1), 2016.

\bibitem{Zhao2017}
S.~Zhao, J.~Song, and S.~Ermon.
\newblock {Learning Hierarchical Features from Generative Models}.
\newblock {\em ICML}, 2017.

\end{thebibliography}
}

\newpage
\appendix
\section{WDC dataset}
\label{sect:datasets}
In this dataset, we consider a more realistic application where we want to find rare events from a large image archives such as satellite images or street view images. To evaluate our method on such applications, we created a new large-scale dataset for detecting newly constructed or destructed buildings in Washington D.C. area from a large archive of aerial images taken in 1995 through 2015.

\subsection{Source data and annotations}
For the WDC dataset, we acquired the aerial images and the building footprints of Washington D.C. area from open data repository hosted by the Government of District of Columbia \cite{WDC}. We used the aerial images taken in 1995, 1999, 2002, 2005, 2008, 2010, 2013 and 2015. The images have 16 cm resolution, and covers over 200$km^2$ for each year. We automatically annotated changes in buildings by comparing the building footprints produced at different years. Specifically, for each building, we computed $IoU$ between footprints of different years, and annotated the buildings as newly constructed or destructed if the $IoU < 0.01$. We conducted the annotation using the footprint pairs of 1999 and 2005, and of 2010 and 2015. Although the building footprints are provided for other years, we find that the difference between them includes many buildings that are missed in the previous year's survey. Because they are not the true change, we decided not to use the footprints.

\subsection{Patch pairs for training and evaluation}
First, we paired off with the aerial images that are close to each other in their acquisition year. From the image pairs, we cropped a large amount of noisy negative patch pairs and a small amount of purely negative and purely positive patch pairs. The purely negative and purely positive patch pairs are created from the change annotations. The noisy negative patch pairs are randomly cropped from all the image pairs available regardless of the existence of the change annotation. Because our targets are rare events, such random samples are almost negative except accidentally cropped positives. The noisy negative samples are used to train our representation learning model. Note that  the training is fully unsupervised because creating noisy negative samples requires no labels. Then, a part of purely negative and purely positive samples are used for fine-tuning the change classifier, and the rest of samples are used for evaluating the model performance.

\subsection{Filtering patches using building footprint}
Since the source aerial images include large areas without buildings, randomly cropped patches rarely contain buildings. In order to put more attention to buildings, we used the building footprint of 2015 as side information to filter out patches that are not containing buildings. Specifically, we controlled the ratio of patches containing and not containing buildings as 9:1.

\section{Hyperparameter settings}
\label{sect:arch}
In the experiments, we used two encoder architectures for the proposed method: one is for Augmented MNIST (Enc-MNIST) and the other is for ABCD, PCD, and WDC datasets (Enc-VGG). Table~\ref{tb:table_arch_vae} shows the detailed architectures of the encoders. The decoders are set as symmetric to the encoders by replacing convolution with deconvolution and subsampling with upsampling. For fair comparisons in terms of network capacity, we basically used the same encoder architectures for the baseline models so that they have the same capacity in fine-tuning phase (see Table~\ref{tb:arch_of_each_model}). As an exception, we needed to use the specific architecture for the model of mathieu et al. \cite{Mathieu2016}, in order to stabilize the GAN training.
Specifically, we modified the Enc-MNIST by inserting instance normalization after every convolution. Moreover, for ABCD, PCD and WDC datasets, we used DCGAN-based architecture.

\begin{table}[tb]
\centering
\caption{Encoder architecture used for Augmented MNIST (Enc-MNIST), and for ABCD, PCD, and WDC dataset (Enc-VGG). The Enc-VGG is based on VGG-16 \cite{Simonyan2014}. Several layers in the encoders have connections to the hidden layers as listed in the 5th column of the table (``Hidden''). For example, ``H1'' in the column represents connection to the operation ``Conv(H1)''. The bottom part of the table shows the architectures of hidden layers, where the number of common and specific features are shown in the second column (``\#Features'')}
\label{tb:table_arch_vae}

\begin{tabular}{c}

\begin{minipage}{0.95\hsize}
\centering
(a) Enc-MNIST
\scalebox{0.65}{
\begin{tabular}{cccccc}
\hline
Operations & \#Features   & Kernel size & Stride & hidden & Spatial dimensions \\ \hline
Conv-ReLU  & 20           & $5\times5$  & 1      & -      & 24                 \\
Max-pool   & 20           & $2\times2$  & 2      & -      & 12                 \\
Conv-ReLU  & 50           & $5\times5$  & 1      & -      & 8                  \\
Max-pool   & 50           & $2\times2$  & 2      & -      & 4                  \\
Conv-ReLU  & 500          & $5\times5$  & 1      & H1     & 1                  \\ \hline
Conv(H1)   & c,s=(20, 10) & $1\times1$  & 1      & -      & 1                  \\ \hline
\end{tabular}
}
\end{minipage} \\
\begin{minipage}{0.95\hsize}
\centering
\vspace{3mm}
(b) Enc-VGG
\scalebox{0.65}{
\begin{tabular}{cccccc}
\hline
Operations & \#Features  & Kernel size  & Stride & Hidden & Spatial dimensions \\ \hline
Conv-ReLU  & 64          & $3\times3$   & 1      & -      & 128                \\
Conv-ReLU  & 64          & $3\times3$   & 1      & H1     & 128                \\
Max-pool   & 64          & $2\times2$   & 2      & -      & 64                 \\
Conv-ReLU  & 128         & $3\times3$   & 1      & -      & 64                 \\
Conv-ReLU  & 128         & $3\times3$   & 1      & H2     & 64                 \\
Max-pool   & 128         & $2\times2$   & 2      & -      & 32                 \\
Conv-ReLU  & 256         & $3\times3$   & 1      & -      & 32                 \\
Conv-ReLU  & 256         & $3\times3$   & 1      & -      & 32                 \\
Conv-ReLU  & 256         & $3\times3$   & 1      & H3     & 32                 \\
Max-pool   & 256         & $2\times2$   & 2      & -      & 16                 \\
Conv-ReLU  & 512         & $3\times3$   & 1      & -      & 16                 \\
Conv-ReLU  & 512         & $3\times3$   & 1      & -      & 16                 \\
Conv-ReLU  & 512         & $3\times3$   & 1      & H4     & 16                 \\
Max-pool   & 512         & $2\times2$   & 2      & -      & 8                  \\
Conv-ReLU  & 1024        & $7\times7$   & 1      & -      & 2                  \\
Conv-ReLU  & 1024        & $1\times1$   & 1      & H5     & 2                  \\ \hline
Conv(H1)   & c,s=(0,32)  & $64\times64$ & 1      & -      & 65                 \\
Conv(H2)   & c,s=(0,32)  & $32\times32$ & 1      & -      & 33                 \\
Conv(H3)   & c,s=(0,32)  & $16\times16$ & 1      & -      & 17                 \\
Conv(H4)   & c,s=(16,16) & $8\times8$   & 1      & -      & 9                  \\
Conv(H5)   & c,s=(16,16) & $1\times1$   & 1      & -      & 2                  \\ \hline
\end{tabular}
}
\end{minipage}

\end{tabular}
\end{table}

\begin{table}[tb]
\centering
\caption{Encoder architectures used for each model.}
\label{tb:arch_of_each_model}
\scalebox{0.65}{
\begin{tabular}{c|cc}
\hline
                        & Aug. MNIST                    & ABCD, PCD, WDC           \\ \hline
Under samp.             & Enc-MNIST                     & Enc-VGG                  \\
Over samp.              & Enc-MNIST                     & Enc-VGG                  \\
Transfer                & Enc-MNIST                     & Enc-VGG                  \\
MLVAE \cite{Bouchacourt2017}                   & Enc-MNIST                     & Enc-VGG, one hidden      \\
Mathieu et al. \cite{Mathieu2016}          & Enc-MNIST with instance norm. & DCGAN based \\
VAE w/o Sim.       & Enc-MNIST                     & Enc-VGG                  \\
VAE w/ Sim. (ours) & Enc-MNIST                     & Enc-VGG                  \\ \hline
\end{tabular}
}
\end{table}

For the experiments on Augmented MNIST, Adam \cite{Kingma2015} was used for optimization with an initial learning rate of 1.0e-3. A coefficient of weight decay term was set to 5.0e-4. All models were trained for 30 epochs using a batch size of 100. $\lambda_{1}$ and $\lambda_{2}$ were both set to 1, and a sparsity parameter of 0.5 was used. 

For the experiments on ABCD, PCD, and WDC datasets, Adam optimizer was used for optimization with an initial learning rate of 1.0e-4. The learning rate decayed linearly with the number of epochs. A coefficient of weight decay term was set to 5.0e-4. All models were trained for 30 epochs using a batch size of 50. $\lambda_{1}$ was set to 10, and $\lambda_{2}$ was set to 1. We also used the KL annealing technique \cite{Bowman2016}, which gradually ascend the coefficient on the KL divergence term in the VAE loss function. The weight was gradually increased during training from 0 to 1.

\section{Ablation Study: kernel size for hidden layers}
\label{sect:ablation}
In Table~\ref{tb:ablation_kernel_size}, we demonstrated a sensitivity analysis on the kernel size of the hidden layers. The result shows that the performance improves when we use larger kernels at the lower layers. This result can be explained as follows: feature maps in lower layers somewhat retain raw information from the input. If we use a small kernel in lower hidden layers, the spatial information is almost all retained, which enables the model to easily reconstruct input images. When the model can reconstruct inputs by using only lower hidden layers, higher layers cannot receive sufficient error signals and the training will be stacked. On the other hand, if we use large kernels, the convolution operation becomes close to fully connected, which promotes abstraction of the hidden variables.

\begin{table}[tb]
\centering
\caption{Sensitivity analysis of the kernel size of hidden layers.}
\label{tb:ablation_kernel_size}
\scalebox{0.9}{
\begin{tabular}{cccccc}
\hline
\multicolumn{5}{c}{Kernel size} &              \\
H1   & H2   & H3   & H4   & H5  & Acc.         \\ \hline
64   & 32   & 16   & 8    & 1   & 90.01(±1.39) \\
32   & 32   & 16   & 8    & 1   & 89.35(±1.86) \\
16   & 16   & 16   & 8    & 1   & 89.36(±1.36) \\
8    & 8    & 8    & 8    & 1   & 88.79(±1.33) \\
3    & 3    & 3    & 3    & 1   & 88.29(±2.26) \\ \hline
\end{tabular}}
\end{table}

\begin{figure*}
\centering
\includegraphics[width=12cm]{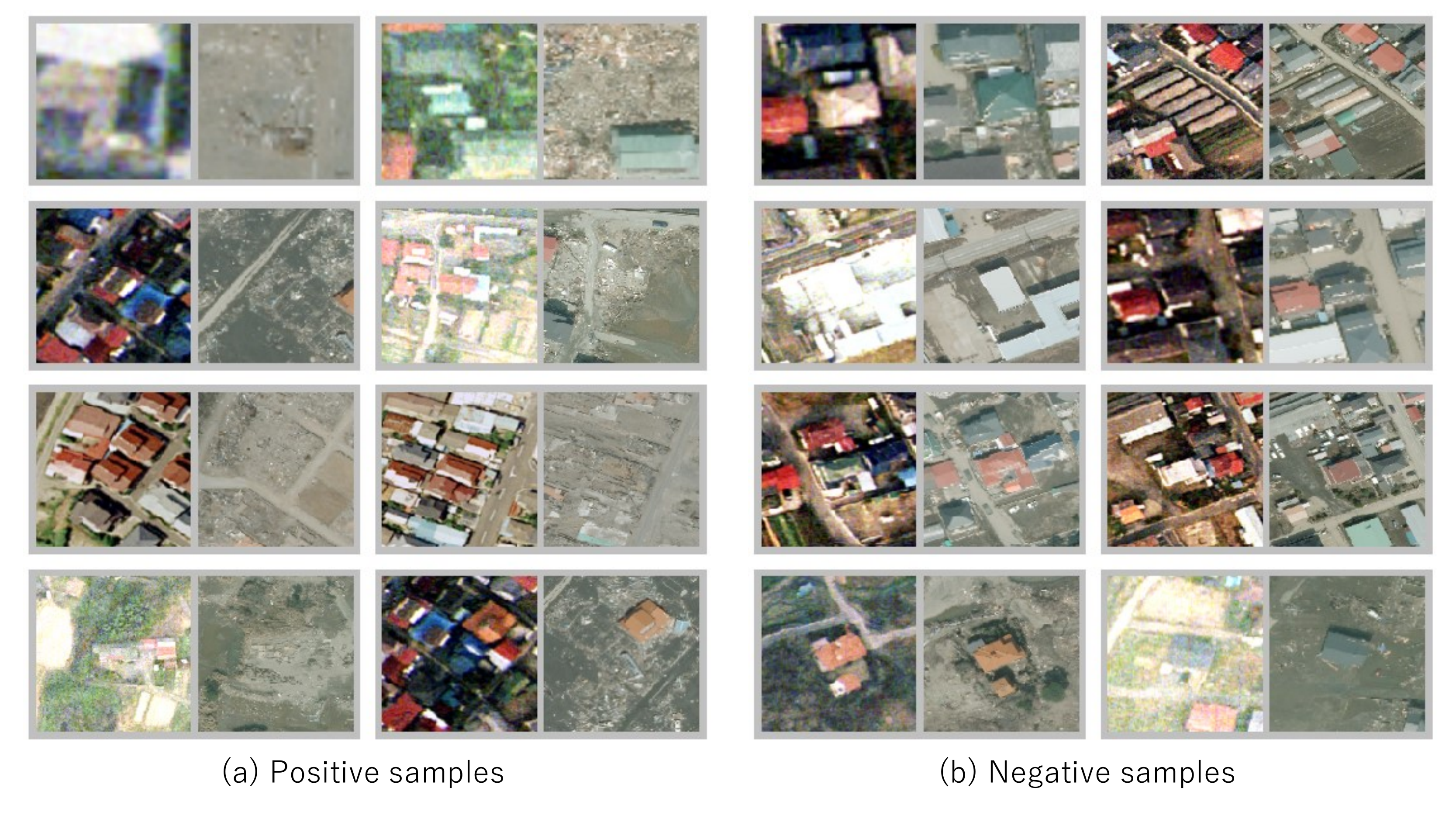}
\caption{Examples of positive pairs and negative pairs in ABCD dataset.}
\label{fig:abcd_samples}
\end{figure*}

\begin{figure*}
\centering
\includegraphics[width=12cm]{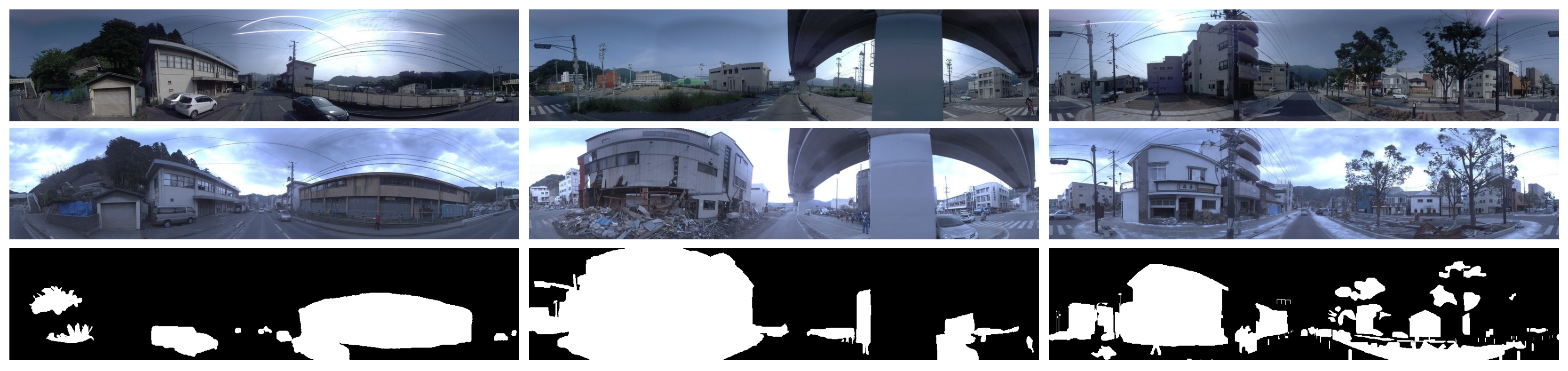}
\caption{Examples of original image pairs in PCD dataset.}
\label{fig:pcd_scenes}
\end{figure*}

\begin{figure*}
\centering
\includegraphics[width=12cm]{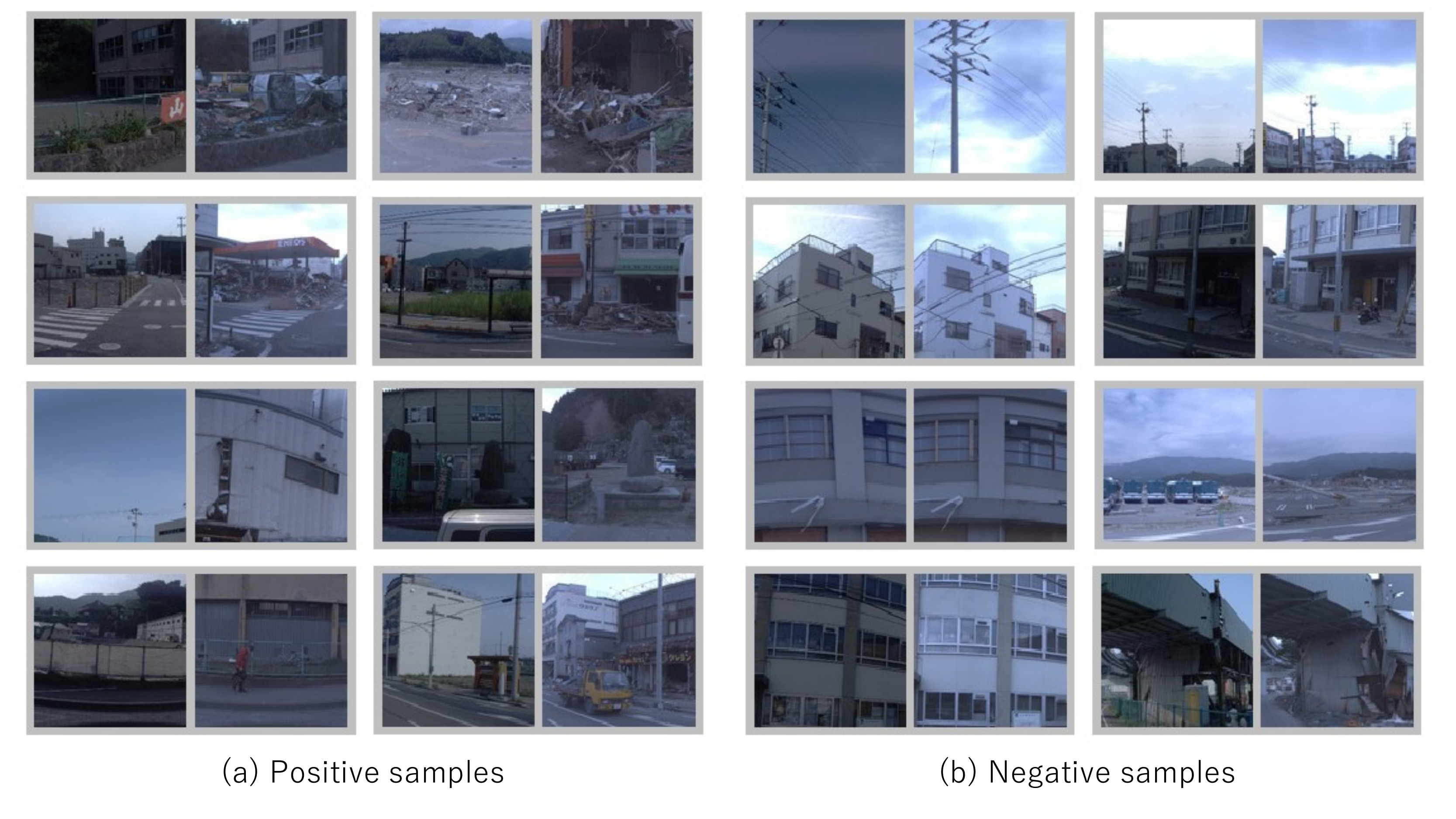}
\caption{Examples of positive pairs and negative pairs cropped from image pairs in PCD dataset.}
\label{fig:pcd_samples}
\end{figure*}

\begin{figure*}
\centering
\includegraphics[width=12cm]{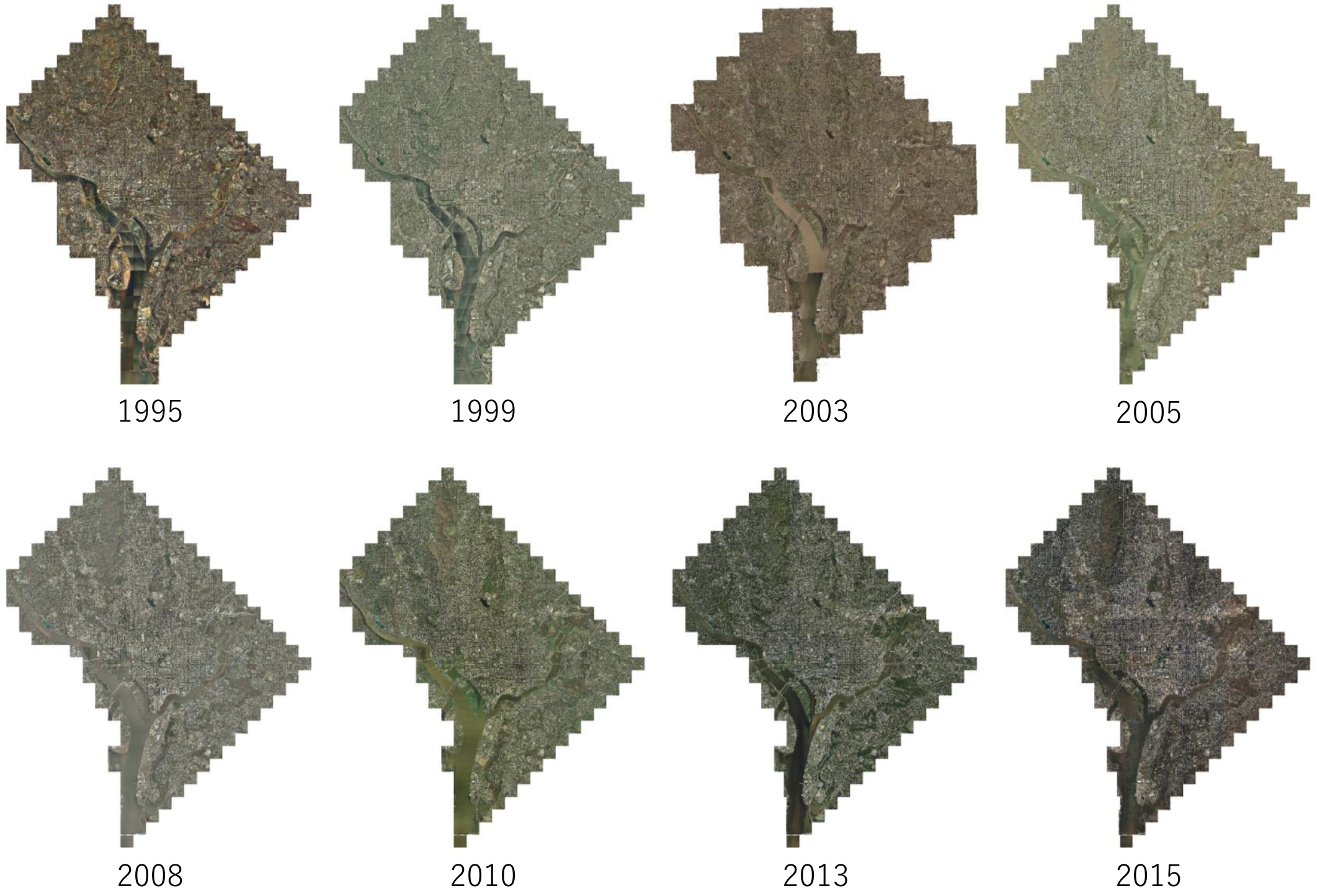}
\caption{Source aerial images of Washington D.C. area.}
\label{fig:wdc_images}
\end{figure*}

\begin{figure*}
\centering
\includegraphics[width=12cm]{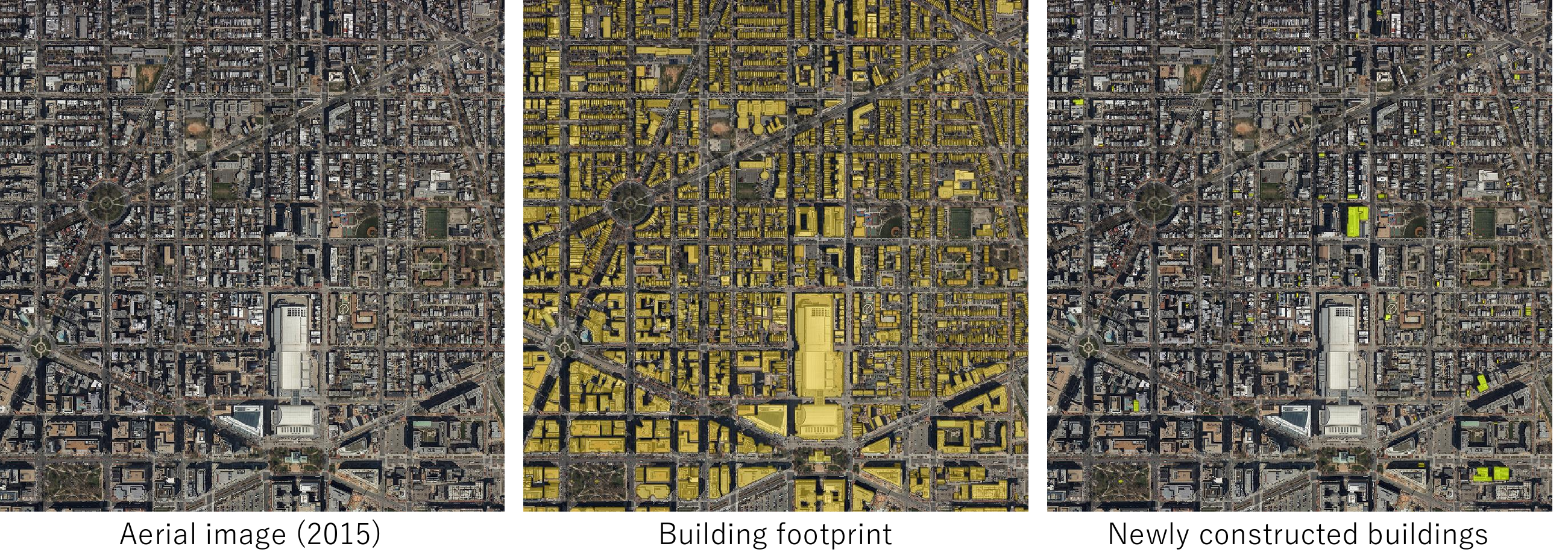}
\caption{Example of aerial images, building footprints and change labels for WDC dataset. The change label on the right is created by comparing building footprints of 2015 and 2010.}
\label{fig:wdc_labels}
\end{figure*}

\begin{figure*}
\centering
\includegraphics[width=12cm]{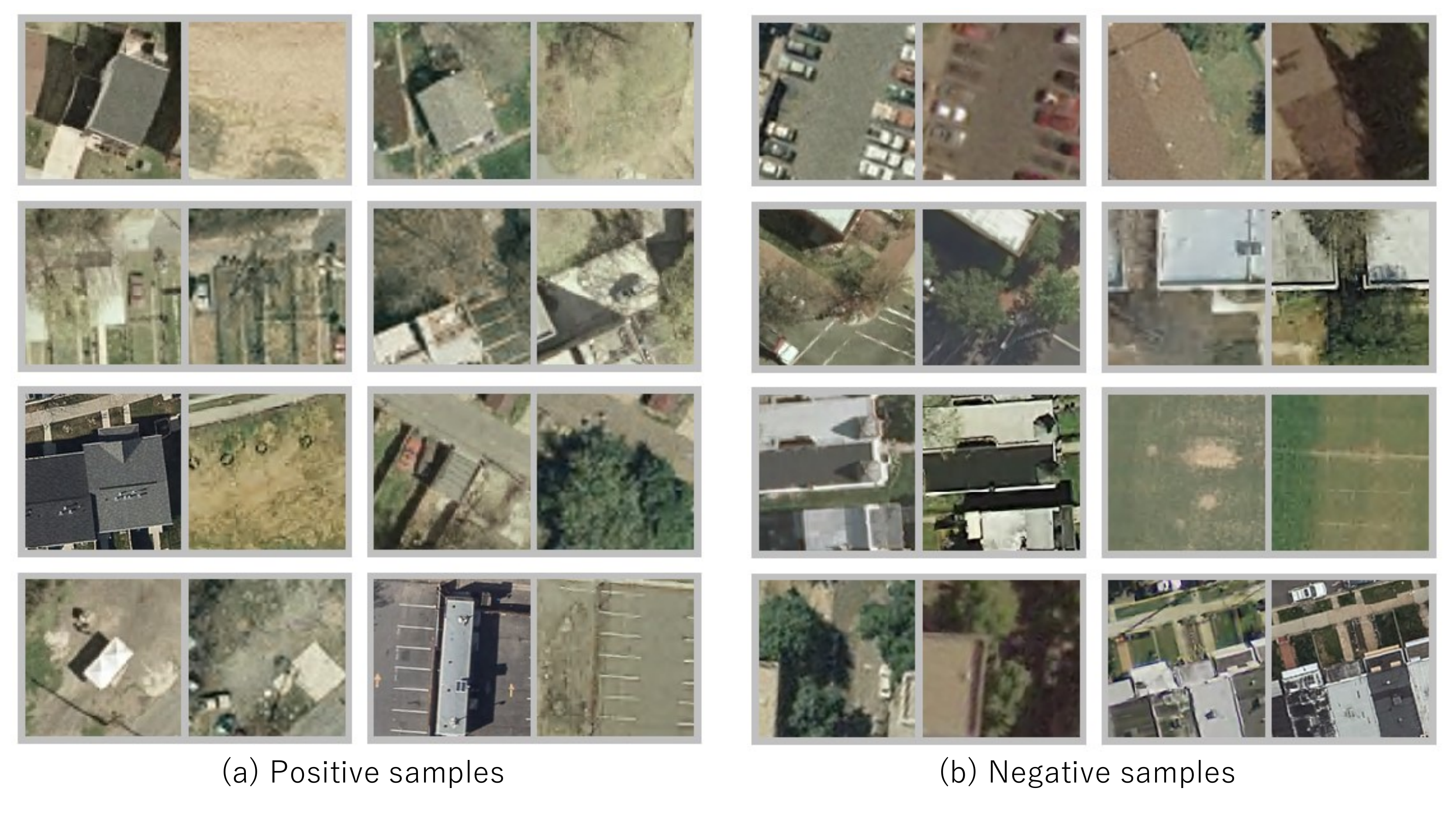}
\caption{Examples of purely positive pairs and noisy negative pairs in WDC dataset.}
\label{fig:wdc_samples}
\end{figure*}

\end{document}